\documentclass{article}

\usepackage{arxiv}
\usepackage[linesnumbered,ruled,vlined]{algorithm2e}
\usepackage[utf8]{inputenc} 
\usepackage[T1]{fontenc}    
\usepackage{hyperref}       
\usepackage{url}            
\usepackage{booktabs}       
\usepackage{amsfonts}       
\usepackage{nicefrac}       
\usepackage{microtype}      
\usepackage{lipsum}
\usepackage{graphicx}
\graphicspath{ {./images/} }
\usepackage{authblk}
\usepackage{amsmath} 

\SetKwInOut{KwParam}{Setting} 
\SetKwInOut{KwIn}{Input}
\SetKwInOut{KwOut}{Output}

\title{Fair Distributed Machine Learning with Imbalanced Data as a Stackelberg Evolutionary Game}

\author[1,2]{Sebastian Niehaus\thanks{Corresponding author: niehaus@cbs.mpg.de}}
\author[2]{Ingo Roeder}
\author[1,3]{Nico Scherf\thanks{NS is supported by BMBF (Federal Ministry of Education and Research) through the Center for Scalable Data Analytics and Artificial Intelligence (ScaDS.AI) and ACONITE (01IS22065)}}

\affil[1]{Max Planck Institute for Human Cognitive and Brain Sciences, Germany}
\affil[2]{Institute for Medical Informatics and Biometry (IMB), TU Dresden, Germany}
\affil[3]{Center for Scalable Data Analytics and Artificial Intelligence (ScaDS.AI), Germany}

\date{} 

\begin{document}
\maketitle
\begin{abstract}
Decentralised learning enables the training of deep learning algorithms without centralising data sets, resulting in benefits such as improved data privacy, operational efficiency and the fostering of data ownership policies. However, significant data imbalances pose a challenge in this framework. Participants with smaller datasets in distributed learning environments often achieve poorer results than participants with larger datasets.\newline
Data imbalances are particularly pronounced in medical fields and are caused by different patient populations, technological inequalities and divergent data collection practices. \newline
In this paper, we consider distributed learning as an Stackelberg evolutionary game. We present two algorithms for setting the weights of each node's contribution to the global model in each training round: the Deterministic Stackelberg Weighting Model (DSWM) and the Adaptive Stackelberg Weighting Model (ASWM). We use three medical datasets to highlight the impact of dynamic weighting on underrepresented nodes in distributed learning. Our results show that the ASWM significantly favours underrepresented nodes by improving their performance by 2.713\% in AUC. Meanwhile, nodes with larger datasets experience only a modest average performance decrease of 0.441\%.
\end{abstract}


\section{Introduction}
Decentralized data refers to the distribution of data across multiple, often geographically dispersed locations or sources, rather than centralizing it at a single site, server, or storage location. This decentralization of data is becoming more common due to the proliferation of connected devices, edge computing, and privacy concerns. While decentralized data offers advantages in terms of data security, privacy, and accessibility, it poses significant challenges for the training of machine learning algorithms. \newline
The challenge of decentralised data is addressed through decentralised machine learning \cite{alsagheer2023decentralized} \cite{bonawitz2017practical}  by enabling model training across multiple nodes without the need to centralise the data. Techniques such as federated learning \cite{konevcny2016federated} allow the data to remain on local devices, while only model updates are shared and aggregated, preserving privacy and reducing the risk of data breaches \cite{zhao2018federated}. This approach not only increases data security, but also enables compliance with data protection regulations and improves scalability by utilising the computing power of numerous decentralised nodes. \newline
A particular challenge in these decentralized learning setups are domains with very different distributions in the individual nodes \cite{hsieh2020non}. This problem is referred to as non-independent and identically distributed (non-iid) data \cite{li2019convergence} and concerns distribution differences in the labels of the data that can arise due to user behaviour, geographical differences, different levels of knowledge, socio-cultural differences or technical differences in the recording devices \cite{luo2021no}. In medical use cases, the problem arises due to the large differences between the nodes that are also the data generators. In medical domains, these are doctor's offices, hospitals and clinics. The reason can basically be divided into these three areas:
\begin{itemize}
\item Changes in patient populations: Patient demographics can shift due to age, gender, and regional factors \cite{feinberg1995prevalence} \cite{cirillo2020sex}. For instance, the proximity of specialist centers or relocation of youth in rural areas can influence the patient base \cite{weinhold2014understanding} \cite{peterson2011rural}. Environmental factors also play a role.
\item Data collection practices: Hospitals vary in their data collection methods and expertise \cite{thakur2020knowledge}, leading to inconsistencies and gaps in the dataset.
\item Technological disparities: Unequal access to modern diagnostic tools among hospitals results in missing data for certain conditions, skewing the overall dataset \cite{kelly2019key} \cite{parikh2019addressing}.
\end{itemize}
These factors create challenges for decentralized machine learning. Each node optimizes the best model for its own patient data. However, the model must also handle unfamiliar cases, as these reflect patients who are not the majority. This makes data from other nodes crucial for improving the model’s performance. The influence of external data depends on factors like data quantity, quality, class distribution, and differences between clinics. Moreover, the imbalance in clinic sizes leads to uneven data contributions in the decentralized learning setup \cite{wu2022node}. \newline
The imbalance in data distribution and data quality in a distributed learning system can be remedied by reweighting the model updates \cite{zhao2024fedsw} \cite{wang2020optimizing}. The updates are reweighted either according to their impact on the global model \cite{wu2021fast} or according to the node-specific loss size. This approach adjusts the impact of each node's updates and ensures that nodes with higher losses (indicating less representative data) contribute more to the global model updates. In this way, the method attempts to mitigate biases caused by uneven data distributions across nodes and promote a more balanced and accurate learning process\cite{reyes21} \cite{deng2022improving} \cite{ZENG2025104994}. The reweighting can be performed statically directly by changing the formula of the federated averaging by using the precision as weighting factor \cite{reyes21} or it can also be learnt dynamically \cite{deng2022improving}. FedVSA enhances traditional averaging by using inference loss to prioritize valuable local updates without revealing data. Its two-step approach detects and mitigates poisoning attacks by analyzing differential inference loss and Shapley values to identify and discard harmful updates \cite{LI2025104990}. Due to the sequential nature of distributed model training, the dynamic reweighting of model updates can also be modelled as a reinforcement learning problem \cite{wang2020optimizing}. In this work, we investigate the weighting and construction of a decentralised machine learning setup using the Stackelberg Game Theory in order to precisely address these imbalances and the associated problems. 
\newline  \newline 
Game theory \cite{vonneumann1944} provides a systematic framework for the analysis of competitive and collaborative games \cite{fudenberg1993game} as distributed learning setups, making it a tool for the development of rules and regulations in the field of distributed learning. By applying game theory principles, we can derive valuable insights and guidelines that can inform future regulations and policies in this rapidly evolving domain. Stackelberg Games \cite{simaan1973stackelberg} are strategic models in game theory where one player, the leader, makes a decision or takes an action. The follower then reacts to this move by making their own decision, optimizing their outcome based on the leader’s choice. The leader anticipates the follower's reaction and chooses his strategy to maximise his own payoff, taking into account the optimal reaction of the follower. This sequential decision-making process is in contrast to games with simultaneous moves, as it allows the leader to exploit his first-mover advantage and influence the follower's actions to his advantage. These games are often used to analyse competitive scenarios in markets, e.g. pricing strategies between a dominant company and smaller competitors \cite{simaan1973additional}. \newline
Stackelberg games allow to explore imbalances in competitive situations  \cite{JIE2018159}. They inherently incorporate asymmetries by designating a leader and a follower, reflecting real-world scenarios where one party often has a strategic advantage or superior information \cite{BENSASSI2024108825}. This framework helps in understanding how imbalances in power, information, or resources affect the interactions and outcomes between competitors \cite{yu2016supply} \cite{zugno2012modeling}. 
\newline \newline
A specialized variant of this framework, known as Stackelberg evolutionary games, extends the concept to dynamic systems involving populations of agents \cite{stein2023stackelberg}. In these games, the leader's strategic decisions shape the evolutionary development of the followers, affecting their decision-making and resulting outcomes over time. This approach captures the dynamics of leader-follower interactions in evolving environments, making it highly relevant to distributed systems and population-based scenarios.
\newline
Recent advances in distributed learning, distributed computing, and game theory have utilized Stackelberg models to address the challenges of participation modeling and control. The task offloading problem, for instance, can be modeled as an evolutionary Stackelberg game \cite{kim2024hierarchical, CHEN202318}, demonstrating notable improvements in energy efficiency and scalability. A hierarchical approach, as presented in \cite{LI2024121023}, integrates resource allocation strategies based on individual client payoffs, further enhancing system performance. Additionally, another model combines dynamic reinforcement learning with game theory to tackle the same problem \cite{DAI2022108}.
\newline
In contrast to these state-of-the-art approaches, the setup in this research intends that every client is included and can dynamically control their participation in the global model. This approach not only maximizes individual rewards and addresses data imbalance issues, ultimately improving the overall model quality. To achieve this, we introduce two algorithms that utilize a contribution weight to control the influence on the global model. The Deterministic Stackelberg Weighting Model (DSWM) uses a deterministic approach to select the contribution weight, while the Adaptive Stackelberg Weighting Model (ASWM) employs a neural network to predict the contribution weight. Both algorithms are validated on three medical datasets in a setup characterized by strong data imbalance and strong class imbalance on each node. They are compared against two baselines: Federated Averaging (FedAvg)\cite{li2019convergence} and Precision Weighted Federated Averaging (PWFedAvg)\cite{reyes21}.

\section{Distributed learning as a Stackelberg Game}\label{sec2}
In this section, we model a distributed learning setup as a three-player Stackelberg game, where each player represents a computation node. Specifically, there is a leader computation node that uses a large proportion of the data for training, and two follower nodes that each have a significantly smaller proportion of the data. 

\subsection*{Stackelberg Evolutionary Game}
A Stackelberg game is a strategic model in which players make decisions sequentially rather than simultaneously. Named after the German economist Heinrich von Stackelberg, this framework is particularly relevant in scenarios where one player (the leader) influences substantial a conflict situation (game) and thus the subsequent decisions of other players (the followers). The leader, by anticipating the followers' responses, selects a strategy that maximizes their own outcome while considering the impact of their decision on the followers' choices. In distributed learning with imbalanced data strong differences of node on the global model, the node with the large datasets is representing the leader and is able to update the global model in a way that other updates do not have any influences

In a Stackelberg game, the leader commits to a strategy that is then observed by the followers. The followers, in turn, choose their strategies in response, aiming to optimize their payoffs given the leader’s decision. The equilibrium reached in such a game is referred to as the Stackelberg equilibrium, wherein neither the leader nor the followers have an incentive to deviate from their chosen strategies.

Stackelberg evolutionary games extend this concept into a dynamic context where players' strategies evolve over time. Rather than making static, one-time decisions, players in evolutionary Stackelberg games continuously adapt and refine their strategies based on iterative interactions and accumulated outcomes. This dynamic evolution can involve mechanisms such as learning, imitation, or adaptation, leading to a continuously evolving landscape of strategies. In distributed learning setups with imbalanced data, where there are strong differences among the nodes in terms of dataset size and class imbalances, the node with the largest dataset effectively acts as the leader. This dominant node can influence the global model's updates so significantly that the contributions from other nodes are underrepresented or even ignored in the final global model. The underrepresented nodes (followers) have to develop a strategy to increase their influence on the global model.

However, distributed learning is better modeled as an evolutionary Stackelberg game due to its inherently dynamic and iterative nature. Unlike traditional Stackelberg games, characterized by a single decision-making event, distributed learning involves ongoing interactions where computing nodes, acting as leaders and followers, repeatedly update their models based on new data and feedback from the system. This iterative process results in a continuous evolution of strategies over time, reflecting the adaptive learning process fundamental to distributed systems. As strategies evolve, they adapt to the changing environment, aligning more closely with the principles of evolutionary Stackelberg games, where the dynamics of learning and adaptation are central to shaping the system's behavior.

\subsection*{Stackelberg Game with Two Followers }
In the context of distributed learning, multiple compute nodes collaborate to train a global model by iteratively updating their local models and incorporating these updates into the global model. This process can be effectively modeled as a Stackelberg game involving three players: a leader (Player 1) and two followers (Players 2 and 3). 

In this constellation, the leader selects a contribution weights for its model updates to optimize its local loss functions. The follower, who observes these weighting factors, then optimizes its own objective function based on the updates it receives. This hierarchical decision-making process reflects the nature of a Stackelberg game, where the leader (Players 1) act first by choosing their strategies (weighting factors) and the followers (Player 2 and 3) responds by optimizing its objective function.

By formulating the distributed learning problem in this way, we can utilize the strategic interactions between the players to improve the overall performance of the global model. The goal of each leader is to influence the update of the global model in a way that minimizes its local loss, while the follower tries to find an optimal response that integrates the contributions of the leader. \newline This means that the follower adjusts its influence on the global model by increasing or decreasing its contribution weights, based on the global model that results from the leader’s update and the previous updates from all players. This approach enables a structured and strategic framework for distributed learning that fosters collaboration and improves model convergence.

Let:
\begin{itemize}
    \item $C_0$ be the contribution weight chosen by the leader (Player 0).
    \item $C_1$ and $C_2$ be the contribution weights chosen by the followers (Player 1 and Player 2), respectively.
    \item $L_0(C_0, C_1, C_2)$ be the loss function for the leader (Player 0) on its local test set.
    \item $L_1(C_0, C_1, C_2)$ and $L_2(C_0, C_1, C_2)$ be the loss functions for the followers (Player 1 and Player 2) on their respective local test sets.
\end{itemize}

The optimization problems for the players are as follows:

\textbf{Leader's Problem (Player 0):}
\begin{equation}
\min_{C_0} L_0(C_0, C_1^*(C_0), C_2^*(C_0))
\end{equation}
where $C_1^*(C_0)$ and $C_2^*(C_0)$ are the optimal responses of the followers given the leader's choice of $C_0$.

\textbf{Follower 1's Problem (Player 1):}
\begin{equation}
\min_{C_1} L_1(C_0, C_1, C_2^*(C_0, C_1))
\end{equation}
where $C_2^*(C_0, C_1)$ is the optimal response of Follower 2 given the choices of the leader and Follower 1.

\textbf{Follower 2's Problem (Player 2):}
\begin{equation}
\min_{C_2} L_2(C_0, C_1^*(C_0, C_2), C_2)
\end{equation}
where $C_1^*(C_0, C_2)$ is the optimal response of Follower 1 given the choices of the leader and Follower 2.

The Stackelberg equilibrium is obtained by solving these problems sequentially, starting with the followers and then proceeding to the leader, taking into account the interdependencies of the strategies.

By formulating the distributed learning problem in this way, we can utilise the strategic interactions between the players to improve the overall performance of the global model. The goal of a leader is to influence the update of the global model in a way that minimises its local loss, while the follower tries to find an optimal response that effectively integrates the contributions of the leaders. This approach enables a structured and strategic framework for distributed learning that fosters collaboration and improves model convergence. This approach can be generalized to a setup with \(n\) leaders and \(m\) followers, where each leader aims to influence the global model update to minimize its own local loss. In this generalized framework, the leaders interact strategically, each trying to optimize their contribution to the global model. The \(m\) followers, in turn, work to find an optimal response that effectively balances and integrates the diverse contributions of the \(n\) leaders.

By extending this strategy to multiple leaders and followers, the approach maintains its core principles of structured and strategic interactions. This not only fosters collaboration among a larger set of participants but also enhances the robustness and convergence of the global model, as the collective influence of multiple leaders and the corresponding adaptive responses of the followers are taken into account.

\section{Methods}
In this section, we describe the algorithm for the distributed learning system as a Stackelberg game using the modulation described in the previous section. We start with the global update and then describe the training algorithm, where we consider two different methods for determining the contribution weights.

\subsection{Global model updates}
In order to perform a global model update, we adapt the federated averaging formula 
\[
w_{t+1} = \frac{\sum_{k=1}^K n_k w_t^k}{\sum_{k=1}^K n_k}
\]
where \(w_{t+1}\) is the updated global model at round \(t+1\), \(K\) is the total number of participating node, \(n_k\) is the number of data points for node \(k\), and \(w_t^k\) is the model update from node \(k\) at round \(t\). We adapt this formula by replacing \(n_k\) by \(C_k\), which is the selected contribution weight for the computation node \(k\). This results in a global model update of the form 
\[
w_{t+1} = \frac{\sum_{k=1}^K C_k w_t^k}{\sum_{k=1}^K C_k}
\]

\subsection{Model update weighting as a Stackelberg game}

This section describes the methodologies employed in the development and evaluation of two distinct approaches for optimizing model contribution weights within a Stackelberg framework: the Deterministic Stackelberg Weighting Model (DSWM) and the Adaptive Stackelberg Weighting Model (ASWM).

The DSWM leverages a fixed, predefined set of contribution weights, denoted as \( C = \{0.1, 0.2, \ldots, 1.0\} \). The objective in DSWM is to select the weight \( c \) from this set that minimizes the loss of the leader or the followers. This selection of the contribution weight requires that the loss for each candidate for the contribution weight is calculated and the contribution weight that yields the lowest loss is selected.

\begin{algorithm}[htbp]
\caption{ASWM Training - Model Update Weighting as a Stackelberg Game}
\label{alg:aswm_training}
\KwIn{Global model $W_{global}$, Initial contribution weights $C^{(0)} = \{c_{F_1}^{(0)}, c_{F_2}^{(0)}\}$, Local datasets $\mathcal{D}_L$, $\mathcal{D}_{F_1}$, $\mathcal{D}_{F_2}$}
\KwParam{Learning rates for leader and followers, Number of training rounds $T$}
\KwOut{Updated global model $W_{global}$, Optimal contribution weights $C^* = \{c_{F_1}^*, c_{F_2}^*\}$}

Initialize $W_{global}$ randomly\;
Initialize contribution weights $C^{(0)} = \{c_{F_1}^{(0)}, c_{F_2}^{(0)}\}$\;
Initialize experience replay buffer\;

\For{each round $t$ in $1$ to $T$}{
    \textbf{Leader's Step (Leader Node):}\;
    Predict the optimal contribution weight $c_{F_1}$ and $c_{F_2}$ using the leader's neural network $\mathcal{N}_L$\;
    Update the leader's local model $W_L$ using $c_{F_1}$ and $c_{F_2}$\;
    Compute the leader's loss $U_L = \mathcal{L}_L(W_L; \mathcal{D}_L)$\;
    Store the predicted $C = \{c_{F_1}, c_{F_2}\}$ and resulting loss $U_L$ in the experience replay buffer\;

    \textbf{Followers' Steps (Follower Nodes):}\;
    \For{each follower $i$ in $\{F_1, F_2\}$}{
        Observe the leader's contribution weights $C$ and loss $U_L$\;
        Predict the follower's optimal contribution weight $c_{F_i}$ using the shared followers' neural network $\mathcal{N}_F$\;
        Update the follower's local model $W_{F_i}$ using $c_{F_i}$\;
        Compute the follower's loss $U_{F_i} = \mathcal{L}_{F_i}(W_{F_i}; \mathcal{D}_{F_i})$\;
        Store the predicted $c_{F_i}$ and resulting loss $U_{F_i}$ in the experience replay buffer\;
    }

    \textbf{Experience Replay:}\;
    Sample past experiences from the replay buffer\;
    Train the leader's neural network $\mathcal{N}_L$ and the shared followers' neural network $\mathcal{N}_F$ using the sampled experiences\;
    Update contribution weights $C = \{c_{F_1}, c_{F_2}\}$ based on the new knowledge\;

    \textbf{Global Model Update:}\;
    Aggregate the updated local models $W_L$, $W_{F_1}$, $W_{F_2}$ using the current contribution weights $C = \{c_{F_1}, c_{F_2}\}$\;
    Update the global model $W_{global}$\;
}

\Return{$W_{global}$, $C^* = \{c_{F_1}^*, c_{F_2}^*\}$}\;
\end{algorithm}

To predict the reactions of the followers, the leader anticipates and accounts for these responses by utilizing an experience replay mechanism \cite{Fedus2020}. This mechanism involves storing and reusing past experiences of follower reactions to different contribution weights. By incorporating these historical reactions, the leader can make more informed decisions, selecting the weight \( c \) that best aligns with minimizing the cumulative loss across all nodes.

In practice, for each contribution weight \( c \) in the set, the leader applies this weight to the model updates and then evaluates the resulting performance on a validation set. The weight that yields the lowest validation loss is chosen as the optimal contribution weight for the current iteration.

The Adaptive Stackelberg Weighting Model (ASWM) builds upon the deterministic approach by introducing a dynamic, model-driven method for setting the contribution weights. Instead of selecting from a predefined set of weights, ASWM trains a model to predict the optimal contribution weight \( c_{\pi} \) based on the current state of the system. 

The model \( W_{\pi} \), which predicts the contribution weight, is trained iteratively using actor-critic, where the DSWM represents the baseline (Training is shown in Algorithm \ref{alg:aswm_training}). The model aims to learn a policy that maximizes overall performance by adjusting the contribution weight in response to changes in the system's state. Similar to DSWM, ASWM uses experience replay to simulate the expected reactions of the followers to different contribution weights. This replay mechanism allows the leader to anticipate how changes in the contribution weight will affect the behavior of the followers and the overall system loss.

In the Stackelberg game setup, we can mathematically represent the neural network’s inputs and structure as follows:

1. \textbf{Leader's Neural Network:}  
   The leader node has its own neural network, denoted as \( \mathcal{N}_L \), which predicts its strategies and interacts with the global model:
   \[
   \mathcal{N}_L(X_L; \theta_L)
   \]
   where:
   - \( X_L \) is the input vector for the leader's neural network, including its local data and contextual variables.
   - \( \theta_L \) represents the weights of the leader's neural network.

2. \textbf{Followers' Neural Networks:}  
   Each follower node also has its own neural network, but these networks share the same weights across the followers. Let \( \mathcal{N}_F \) denote the shared neural network for followers:
   \[
   \mathcal{N}_F(X_{F_i}; \theta_F)
   \]
   where:
   - \( X_{F_i} \) is the input vector for follower \( i \), which includes local data, context, and previous iteration information.
   - \( \theta_F \) represents the shared weights of the followers' neural networks.

3. \textbf{Shared Follower Weights:}  
   The shared weights \( \theta_F \) are updated collaboratively based on aggregated gradients from all follower nodes. Each follower \( i \) computes gradients locally on its dataset \( \mathcal{D}_{F_i} \), and these are averaged during training:
   \[
   \theta_F^{(t)} = \theta_F^{(t-1)} - \eta \cdot \frac{1}{n_F} \sum_{i=1}^{n_F} \nabla \mathcal{L}_{F_i}(\theta_F; \mathcal{D}_{F_i})
   \]
   where:
   - \( n_F \) is the number of followers.
   - \( \eta \) is the learning rate.
   - \( \mathcal{L}_{F_i} \) is the loss function of follower \( i \), evaluated on its local test set.

4. \textbf{Input Vector for Neural Networks:}
   The input vector for the leader's neural network \( \mathcal{N}_L \) is:
   \[
   X_L = \begin{bmatrix} x_L \\ C^{(t-1)} \\ W^{(t-1)} \\ U_L(x_L, x_{F_1}, x_{F_2}) \end{bmatrix}
   \]
   Similarly, the input vector for the followers' neural network \( \mathcal{N}_F \) is:
   \[
   X_{F_i} = \begin{bmatrix} x_{F_i} \\ C^{(t-1)} \\ W^{(t-1)} \\ U_{F_i}(x_L, x_{F_1}, x_{F_2}) \end{bmatrix}
   \]
   These input vectors incorporate local strategies, previous contribution weights \( C^{(t-1)} \), previous iteration weights \( W^{(t-1)} \), and utility functions \( U_L \) or \( U_{F_i} \) as loss functions on local test sets.

5. \textbf{Output of Neural Networks:}
   - The leader’s neural network outputs the updated contribution weights for the followers:
     \[
     \mathcal{N}_L(X_L; \theta_L) \rightarrow C = \begin{bmatrix} c_{F_1} \\ c_{F_2} \end{bmatrix}
     \]
   - The shared followers’ neural network predicts the local strategies of each follower:
     \[
     \mathcal{N}_F(X_{F_i}; \theta_F) \rightarrow x_{F_i}
     \]

6. \textbf{Global Collaboration:}
   The global model integrates contributions from all nodes (leader and followers) to update its parameters. The contribution weights \( C \) control the influence of each follower on the global model:
   \[
   \theta_G^{(t)} = \theta_G^{(t-1)} + \sum_{i=1}^{n_F} c_{F_i} \cdot \nabla \mathcal{L}_{F_i}(\theta_G; \mathcal{D}_{F_i})
   \]

7. \textbf{Prediction Objective:}
   The leader’s neural network predicts \( C \), the contribution weights controlling each follower’s influence on the global model:
   \[
   C = \mathcal{N}_L(X_L; \theta_L)
   \]
   Each follower’s shared neural network predicts its local strategy:
   \[
   C_{F} = \mathcal{N}_F(X_{F}; \theta_F)
   \]

The training process involves repeatedly updating the model \( W_{\pi} \) based on the observed performance outcomes, gradually refining the ability to select weights that minimize the loss across for the respective player. As the model becomes more proficient, it adapts to the dynamics of the system, allowing for a more responsive and optimized contribution weight selection compared to the deterministic approach.

Both DSWM and ASWM aim to optimize the contribution weights within a Stackelberg framework by accounting for the interaction between the leader and followers. While DSWM relies on a fixed set of weights and selects the optimal one through loss minimization, ASWM uses a trained model to dynamically adapt the contribution weights in response to the system's current state. The use of experience replay in both methods enhances the leader's ability to anticipate follower reactions, ultimately leading to improved performance in distributed learning environments.

\subsection{Experimental setup}
\begin{figure*}[t]
\centering
\includegraphics[width=1\textwidth]{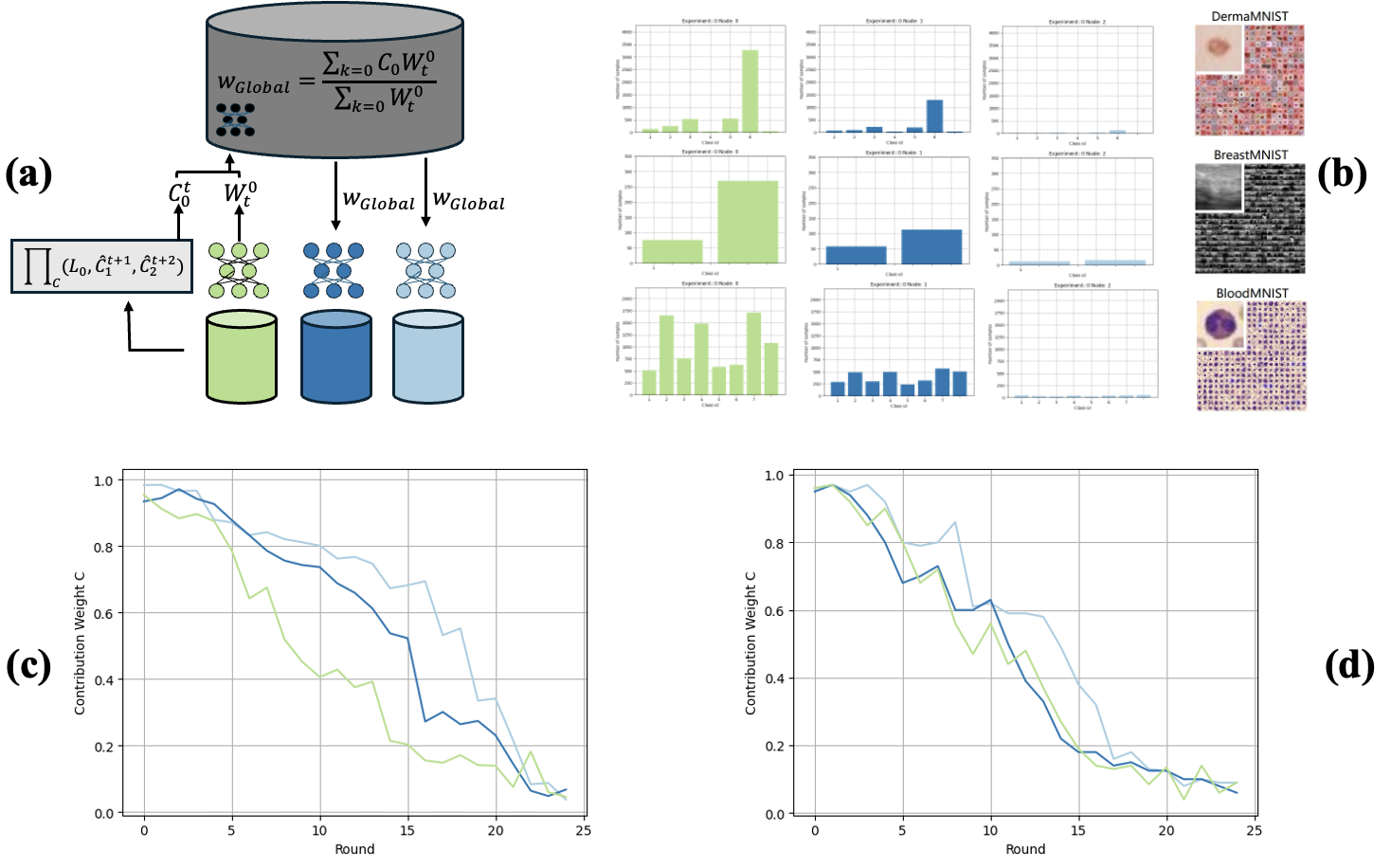} 
\caption{This figure provides a detailed overview of the estimation of contribution weights for model updates and the data distribution across nodes, with the leader represented in light green, the first follower in dark blue, and the second follower in light blue. Panel (a) depicts the leader facilitating a global update. Panel (b) illustrates the class-wise distribution across all datasets and nodes. Panels (c) and (d) present the selection of contribution weights for the respective nodes using the ASWM method in panel (c) and the DSWM method in panel (d). The x-axis is showing the rounds of global model updates.}
\label{fig1}
\end{figure*}

In order to demonstrate the viability of our proposed approach, we conducted experiments using simple datasets that reflect key characteristics of medical imaging data. The MedMNIST datasets, in particular, provide a relevant and manageable starting point for validating the core principles of our method.
The experimental setup involved simulating a distributed machine learning environment across three distinct nodes, each representing a subset of one of the three datasets: BreastMNIST, DermaMNIST, and BloodMNIST (Figure \ref{fig1}). The class distributions over these nodes were generated using a Dirichlet distribution model \cite{lin2016dirichlet}. The Dirichlet distribution is particularly well-suited for modeling unbalanced data, as it represents probabilities across multiple outcomes in inherently skewed scenarios. This model's flexibility in handling varying concentrations captures the relative proportions of different categories, enabling the synthesis of datasets with underrepresented classes.

Each node in the simulation was assigned its own unique Gaussian noise pattern, ensuring that the data noise characteristics varied across the different nodes. This noise addition simulates real-world scenarios where data collected from different sources may have inherent inconsistencies due to varying conditions and environments. As shown in Table \ref{tab:client_distribution}, the probability distribution of data samples among the nodes of the Leader, Follower 1, and Follower 2 varies across the BloodMNIST, DermaMNIST, and BreastMNIST datasets, reflecting these diverse data characteristics.

\begin{table}[h!]
\centering
\begin{tabular}{|c|c|c|c|}
\hline
\textbf{Dataset} & \textbf{Leader} & \textbf{Follower 1} & \textbf{Follower 2} \\ \hline
BloodMNIST       & 0.61            & 0.25                & 0.14                \\ \hline
DermaMNIST       & 0.62            & 0.23                & 0.15                \\ \hline
BreastMNIST      & 0.66            & 0.23                & 0.11                \\ \hline
\end{tabular}
\caption{Averaged probability distribution of data samples among nodes for different datasets}
\label{tab:client_distribution}
\end{table}

The experiments were repeated 10 times to ensure the robustness of the results. This repetition allowed for accounting for the variability introduced by the Dirichlet distribution and Gaussian noise, providing a more reliable assessment of the model's performance across the distributed nodes. The repeated experiments enabled the calculation of average performance metrics, thereby reducing the impact of any outliers or anomalies in the data.

\section{Results}
The results of the study are summarized in terms of the performance metrics, specifically the Area Under the Curve (AUC), for different methods across three nodes for each dataset: BreastMNIST, DermaMNIST, and BloodMNIST. The methods compared include FedAvg, PWFedAvg, the Deterministic Stackelberg Weighting Model (DSWM), and the Adaptive Stackelberg Weighting Model (ASWM).

For the BreastMNIST dataset, the FedAvg method resulted in AUC values of 0.812 for Node 1, 0.729 for Node 2, and 0.614 for Node 3. When the PWFedAvg method was applied, the AUC values improved slightly, with Node 1 reaching 0.823, Node 2 at 0.745, and Node 3 at 0.666. The DSWM method showed further improvement, with Node 1 achieving an AUC of 0.816, Node 2 at 0.768, and Node 3 at 0.671. The ASWM method provided the best performance for this dataset, where Node 1 maintained an AUC of 0.812, Node 2 improved to 0.772, and Node 3 reached 0.674. When comparing ASWM to PWFedAvg, Node 2 saw a percentage improvement of approximately 3.62\%, and Node 3 saw an improvement of approximately 1.20\%. Node 1, however, experienced a slight decrease of about 1.34\%.

For the DermaMNIST dataset, the FedAvg method showed AUC values of 0.803 for Node 1, 0.732 for Node 2, and 0.572 for Node 3. The PWFedAvg method resulted in higher AUC values, with Node 1 at 0.826, Node 2 at 0.745, and Node 3 at 0.553. The DSWM method further increased these values, with Node 1 at 0.824, Node 2 at 0.751, and Node 3 at 0.567. The ASWM method yielded the highest AUC values, with Node 1 at 0.821, Node 2 at 0.769, and Node 3 improving to 0.586. Comparing ASWM to PWFedAvg, Node 2 saw a percentage improvement of approximately 3.22\%, and Node 3 improved by 5.97\%, while Node 1 experienced a slight decrease of about 0.61\%.

In the case of the BloodMNIST dataset, the FedAvg method demonstrated strong performance across all nodes, with Node 1 achieving an AUC of 0.952, Node 2 at 0.944, and Node 3 at 0.942. The PWFedAvg method provided slight improvements, with Node 1 at 0.971, Node 2 at 0.968, and Node 3 slightly lower at 0.939. The DSWM method continued this trend, resulting in Node 1 reaching an AUC of 0.978, Node 2 at 0.964, and Node 3 at 0.951. Finally, the ASWM method achieved the highest AUC values across all nodes, with Node 1 at 0.977, Node 2 at 0.969, and Node 3 at 0.958. In this dataset, the percentage differences between ASWM and PWFedAvg were marginal, with Node 1 decreasing slightly by about 0.10\%, Node 2 improving by 0.10\%, and Node 3 improving by approximately 2.02\%.

Overall, the results demonstrate that the Adaptive Stackelberg Weighting Model (ASWM) consistently outperforms the other methods, particularly for underrepresented nodes (Nodes 2 and 3), across all datasets. The AUC improvement is most pronounced for Node 3, which typically has the smallest datasets, highlighting the effectiveness of the ASWM in addressing data imbalance issues in distributed machine learning setups. While nodes with larger datasets (e.g., Node 1) experience modest gains or maintain performance, the overall benefit is the enhanced equity in model performance across nodes with varying data sizes. The percentage improvements of ASWM over PWFedAvg emphasize its capability in enhancing model performance, especially in scenarios where data is imbalanced.

\begin{table}[h!]
\centering
\begin{tabular}{cccccc}

\textbf{Method} & \textbf{Metric} & \textbf{Leader} & \textbf{Follower 1} & \textbf{Follower 2} \\

FedAvg & AUC & 0.812 & 0.729 & 0.614 \\
PWFedAvg & AUC & \textbf{0.823} & 0.745 & 0.666 \\
DSWM & AUC & 0.816 & 0.768 & 0.671 \\
ASWM & AUC & 0.812 & \textbf{0.772} & \textbf{0.674} \\

\end{tabular}
\caption{Performance metrics for BreastMNIST dataset across different methods and nodes}
\label{tab:breastmnist}
\end{table}

\begin{table}[h!]
\centering
\begin{tabular}{cccccc}

\textbf{Method} & \textbf{Metric} & \textbf{Leader} & \textbf{Follower 1} & \textbf{Follower 2} \\
FedAvg & AUC & 0.803 & 0.732 & 0.572 \\
PWFedAvg & AUC & \textbf{0.826} & 0.745 & 0.553 \\
DSWM & AUC & 0.824 & 0.751 & 0.567 \\
ASWM & AUC & 0.821 & \textbf{0.769} & \textbf{0.586} \\
\end{tabular}
\caption{Performance metrics for DermaMNIST dataset across different methods and nodes}
\label{tab:dermaminst}
\end{table}

\begin{table}[h!]
\centering
\begin{tabular}{cccccc}
\textbf{Method} & \textbf{Metric} & \textbf{Leader} & \textbf{Follower 1} & \textbf{Follower 2} \\
FedAvg & AUC & 0.952 & 0.944 & 0.942 \\
PWFedAvg & AUC & 0.971 & 0.968 & 0.939 \\
DSWM & AUC & \textbf{0.978} & 0.964 & 0.951 \\
ASWM & AUC & 0.977 & \textbf{0.969} & \textbf{0.958} \\
\end{tabular}
\caption{Performance metrics for BloodMNIST dataset across different methods and nodes}
\label{tab:bloodmnist}
\end{table}

\section{Conclusion}
In this work, we explored the challenges of decentralized machine learning with imbalanced data by framing the problem as a Stackelberg evolutionary game. We introduced two algorithms, the Deterministic Stackelberg Weighting Model (DSWM) and the Adaptive Stackelberg Weighting Model (ASWM), designed to dynamically adjust the contribution of each node based on their data characteristics. The results demonstrate that the ASWM consistently improves the performance of underrepresented nodes, leading to a more equitable distribution of learning outcomes across different nodes with varying data sizes.

The ASWM model showed gains in performance, particularly for nodes with smaller datasets, addressing the imbalance issue inherent in decentralized learning setups. This highlights the potential of game theory-based approaches in enhancing the fairness and effectiveness of distributed machine learning systems.

However, a key limitation of our approach is the sequential nature of the Stackelberg game model, which contrasts with the simultaneous updates typical in traditional decentralized learning setups, such as federated learning. This sequential setup, while useful for modeling strategic interactions, may not fully capture the parallelism and efficiency of standard distributed learning frameworks. Future research should investigate how the principles of Stackelberg games can be integrated with more conventional, simultaneous update models to combine the benefits of both approaches. Additionally, further exploration is needed to test the scalability and applicability of these models in more complex and larger-scale decentralized environments.

In conclusion, while the Stackelberg game-based models offer a promising avenue for addressing data imbalance in decentralized learning, the sequential decision-making process represents a departure from typical distributed learning paradigms, and its implications on system performance and scalability warrant further investigation. In future work, the relationship between imbalance and performance should be investigated to better understand how these factors interact and influence the effectiveness of such models.

The objective of this work is to serve as a inspiration for further research aimed at testing and expanding upon the theorems derived from game theory. We hope to inspire researchers to explore and validate the applicability of these game theory concepts to real-world scenarios within distributed learning. By pushing the boundaries of game theory's application, we might shape the future of distributed learning and ensure that it remains fair, efficient, and secure.

\bibliographystyle{plain} 
\bibliography{references}

\end{document}